%% file: acl2019.tex
\documentclass[11pt,a4paper]{article}
\usepackage[hyperref]{acl2019}
\usepackage{times}
\usepackage{latexsym}

\usepackage{url}
\usepackage{enumitem}
\usepackage{pifont}
\usepackage{textcomp}
\usepackage{graphicx}
\usepackage{subfigure}
\usepackage[font={footnotesize}]{caption}
\usepackage{amsmath,amssymb}

\usepackage{varwidth}
\usepackage{booktabs}

\usepackage{url}
\usepackage{arydshln}

\newcommand{\multiqa}{\textsc{MultiQA}}
\newcommand{\triviaqa}{\textsc{TriviaQA}}
\newcommand{\triviaqag}{\textsc{TQA-G}}
\newcommand{\triviaqawiki}{\textsc{TQA-W}}
\newcommand{\triviaqaunfilt}{\textsc{TQA-U}}
\newcommand{\hotpotqa}{\textsc{HotpotQA}}
\newcommand{\squad}{\textsc{SQuAD}}
\newcommand{\newsqa}{\textsc{NewsQA}}
\newcommand{\searchqa}{\textsc{SearchQA}}
\newcommand{\drop}{\textsc{DROP}}
\newcommand{\cwq}{\textsc{CWQ}}
\newcommand{\cq}{\textsc{CQ}}
\newcommand{\comqa}{\textsc{ComQA}}
\newcommand{\wikihop}{\textsc{WikiHop}}
\newcommand{\mixs}{\textsc{Multi-37K}}
\newcommand{\mixa}{\textsc{Multi-75K}}
\newcommand{\mixb}{\textsc{Multi-150K}}
\newcommand{\mixc}{\textsc{Multi-250K}}
\newcommand{\mixd}{\textsc{Multi-300K}}
\newcommand{\mixe}{\textsc{Multi-375K}}
\newcommand{\triviaqamix}{\textsc{AllContexts}}
\newcommand{\self}{\textsc{Self}}

\newcommand{\sotasnum}{five}

\newcommand{\docqa}{\textsc{DocQA}}
\newcommand{\bertqa}{\textsc{BertQA}}

\aclfinalcopy 
\def\aclpaperid{1073} 

\newcommand{\emailsize}{\fontsize{12pt}\times}

\title{\multiqa: An Empirical Investigation of Generalization and Transfer in Reading Comprehension}

\author{Alon Talmor$^{1,2}$  ~~~~~Jonathan Berant$^{1,2}$ \\
\mbox{}\\
$^1$School of Computer Science, Tel-Aviv University \\
$^2$Allen Institute for Artificial Intelligence \\
\emailsize{\texttt{\{alontalmor@mail,joberant@cs\}.tau.ac.il}}}
\date{}  
  
\date{}

\begin{document}
\maketitle

\input{00_abstract}
\input{01_intro}
\input{02_datasets}
\input{03_models}
\input{04_experiments}

\input{05_full_experiments}

\input{06_related}
\input{07_conclusions}

\section*{Acknowledgments}
We thank the anonymous reviewers for their constructive feedback. This work was completed in partial fulfillment for the PhD degree of Alon Talmor. This research was partially supported by The Israel Science Foundation grant 942/16, The Blavatnik Computer Science Research Fund and The Yandex Initiative for Machine Learning.
\bibliography{all}
\bibliographystyle{acl_natbib}

\appendix



\end{document}

%% file: 00_abstract.tex
\begin{abstract}
A large number of reading comprehension (RC) datasets has been created recently, but little analysis has been done on whether they generalize to one another, and the extent to which existing datasets can be leveraged for improving performance on new ones. In this paper, we conduct such an investigation over ten RC datasets, training on one or more \emph{source} RC datasets, and evaluating generalization, as well as transfer to a target RC dataset.
We analyze the factors that contribute to generalization, and show that training on a source RC dataset and transferring to  a target dataset substantially improves performance, even
in the presence of powerful contextual representations from BERT \cite{devlin2019bert}.
We also find that training on multiple source RC datasets leads to robust generalization and transfer, and can reduce the cost of example collection for a new RC dataset. Following our analysis, we propose \multiqa{}, a BERT-based model, trained on multiple RC datasets, which leads to state-of-the-art performance on \sotasnum{} RC datasets. We share our infrastructure for the benefit of the research community.
\end{abstract}

%% file: 01_intro.tex
\section{Introduction}
\label{sec:intro}

Reading comprehension (RC) is concerned with reading a piece of text and answering questions about it \cite{richardson2013mctest,berant2014biological,hermann2015read,rajpurkar2016squad}. Its appeal stems both from the clear application it proposes, but also from the fact that it allows to probe many aspects of language understanding, simply by posing questions on a text document. Indeed, this has led to the creation of a large number of RC datasets in recent years.

While each RC dataset has a different focus, there is still substantial overlap in the abilities required to answer questions across these datasets.
Nevertheless, there has been relatively little  work \cite{min2017transfer,chung2018supervised,sun2018improving} 
that explores the relations between the different datasets, including whether a model trained on one dataset generalizes to another.
This research gap is highlighted by the increasing interest in developing and evaluating the generalization of language understanding models to new setups
\cite{yogatama2019learning,liu2019multi}.

In this work, we conduct a thorough empirical analysis of generalization and transfer across 10 RC benchmarks.
We train models on one or more source RC datasets, and then evaluate their performance on a target test set, either without any additional target training examples (\emph{generalization}) or with additional target examples (\emph{transfer}). We experiment with \docqa{} \cite{clark2018simple}, a standard and popular RC model, as well as a model based on BERT \cite{devlin2019bert}, which provides powerful contextual representations.

Our generalization analysis confirms findings that current models over-fit to the particular training set and generalize poorly even to similar datasets. Moreover, BERT representations substantially improve generalization. However, we find that the contribution of BERT is much more pronounced on Wikipedia (which BERT was trained on) and Newswire, but quite moderate when documents are taken from web snippets. 

We also analyze the main causes for poor generalization: (a) differences in the language of the text document, (b) differences in the language of the question, and (c) the type of language phenomenon that the dataset explores. We show how generalization is related to these factors (Figure~\ref{fig:spring-graph}) and that performance drops as more of these factors accumulate.

Our transfer experiments show that pre-training on one or more source RC datasets substantially improves performance when fine-tuning on a target dataset. 
An interesting question is whether such pre-training improves performance even in the presence of powerful language representations from BERT. We find the answer is a conclusive yes, as we obtain consistent improvements in our BERT-based RC model.

We find that training on multiple source RC datasets is effective for both generalization and transfer. In fact, training on multiple datasets leads to the same performance as training from the target dataset alone, but with roughly three times fewer examples. Moreover, we find that when using the high capacity BERT-large, one can train a single model on multiple RC datasets, and obtain close to or better than state-of-the-art performance on all of them, without fine-tuning to a particular dataset.

Armed with the above insights, we train a large RC model on multiple RC datasets, termed \multiqa{}.
Our model leads to new state-of-the-art results on \sotasnum{} datasets, suggesting  that in  many  language understanding  tasks the size of the dataset is the  main bottleneck, rather than the model itself.

Last, we have developed infrastructure (on top of AllenNLP \cite{gardner2018allennlp}), where experimenting with  multiple models on multiple RC datasets, mixing datasets, and performing fine-tuning, are trivial.
It is also simple to expand the infrastructure to new datasets and new setups (abstractive RC, multi-choice, etc.).
We will open source our infrastructure,
which will help researchers evaluate models on a large number of datasets, and gain insight on the strengths and shortcoming of their methods. We hope this will accelerate progress in language understanding.

To conclude, we perform a thorough investigation of generalization and transfer in reading comprehension over 10 RC datasets. Our findings are:
\begin{itemize}[topsep=0pt, itemsep=0pt, leftmargin=0in, parsep=0pt]
    \item An analysis of generalization on two RC models, illustrating the factors that influence generalization between datasets.
    \item Pre-training on a RC dataset and fine-tuning on a target dataset substantially improves performance even in the presence of contextualized word representations (BERT).
    \item Pre-training on multiple RC datasets improves transfer and generalization and can reduce the cost of example annotation.
    \item A new model, \multiqa{}, that improves state-of-the-art performance on \sotasnum{} datasets. 
    \item Infrastructure for easily performing experiments on multiple RC datasets.
\end{itemize}

The uniform format datasets can be downloaded from  \url{www.tau-nlp.org/multiqa}. The code for the AllenNLP models is available at \url{github.com/alontalmor/multiqa}.

%% file: 02_datasets.tex
\section{Datasets}
\label{sec:dataset}

\begin{table}[t]
\begin{center}
\large{
\resizebox{1.0\columnwidth}{!}{
\begin{tabular}{l|l|l|l|l}
Dataset & Size & Context & Question & Multi-hop  \\
\midrule
\squad & 108K  & Wikipedia & crowd & No  \\
\newsqa & 120K & Newswire & crowd & No  \\
 \searchqa & 140K & Snippets & trivia & No \\
 \triviaqa & 95K & Snippets & trivia & No \\
 \hotpotqa & 113K  & Wikipedia & crowd & Yes \\
 \midrule
 \cq & 2K & Snippets & Web queries/KB & No \\
 \cwq & 35K & Snippets & crowd/KB & Yes \\
 \comqa & 11K & Snippets & WikiAnswers & No\\
 \wikihop & 51K & Wikipedia & KB & Yes  \\
 \drop & 96K & Wikipedia & crowd & Yes  
\end{tabular}}}
\end{center}
\caption{Characterization of different RC datasets. The top part corresponds to \emph{large datasets}, and the bottom to small datasets.} 
\label{tab:datasets}
\end{table}

We describe the 10 datasets used for our investigation. Each dataset provides question-context-answer triples $\{(q_i, c_i, a_i)\}_{i=1}^N$ for training, and a model maps an unseen question-context pair $(q,c)$ to an answer $a$. For simplicity, we focus on the single-turn extractive  setting, where the answer $a$ is a span in the context $c$. Thus, we do not evaluate abstractive \cite{nguyen2016ms} or conversational datasets \cite{choi2018quac,reddy2018coqa}.

We broadly distinguish \emph{large datasets} that include more than 75K examples, from \emph{small datasets} that contain less than 75K examples. In \S\ref{sec:experiments}, we will fix the size of the large datasets to control for size effects, and always train on exactly 75K examples per dataset.

We now shortly describe the datasets, and provide a summary of their characteristics in Table~\ref{tab:datasets}. The table shows the original size of each dataset, the source for the context, how questions were generated, and whether the dataset was specifically designed to probe multi-hop reasoning.

The large datasets used are:
\begin{enumerate}[topsep=0pt,itemsep=0ex,parsep=0ex,leftmargin=*]
\item \squad{} \cite{rajpurkar2016squad}:
Crowdsourcing workers were shown Wikipedia paragraphs and were asked to author questions about their content. 
Questions mostly require soft matching of the language in the question to a local context in the text.
\item \newsqa{} \cite{trischler2017newsqa}:
Crowdsourcing workers were shown a CNN article (longer than \squad{}) and were asked to author questions about its content. 
\item \searchqa{} \cite{dunn2017searchqa}:
Trivia questions were taken from Jeopardy! TV show,
and contexts are web snippets retrieved from Google search engine for those questions, with an average of 50 snippets per question.
\item \textsc{TriviaQA} \cite{joshi2017triviaqa}:
Trivia questions were crawled from the web.
In one variant of \triviaqa{} (termed \triviaqawiki{}), Wikipedia pages related to the questions are provided for each question. In another, web snippets and documents from Bing search engine are given. For the latter variant, we use only the web snippets in this work (and term this \triviaqaunfilt{}). In addition, we replace Bing web snippets with Google web snippets (and term this \triviaqag{}).
\item \hotpotqa{} \cite{yang2018hotpotqa}: 
Crowdsourcing workers were shown pairs of related Wikipedia paragraphs and asked to author questions that require multi-hop reasoning over the paragraphs.
There are two versions of \hotpotqa{}:
the first where the context includes the two gold paragraphs and eight ``distractor" paragraphs, and a second, where 10 paragraphs retrieved by an information retrieval (IR) system are given. Here, we use the latter version.
\end{enumerate}

The small datasets are:
\begin{enumerate}[topsep=0pt,itemsep=0ex,parsep=0ex,leftmargin=*]
\item \cq{} \cite{bao2016constraint}: 
Questions are real Google web queries crawled from Google Suggest, originally constructed for querying the KB Freebase \cite{bollacker2008freebase}. However, the dataset was also used as a RC task with retrieved web snippets \cite{talmor2017evaluating}.
\item \cwq{} \cite{talmor2018web}:
Crowdsourcing workers were shown compositional formal queries against Freebase and were asked to re-phrase them in natural language. Thus, questions require multi-hop reasoning. The original work assumed models contain an IR component, but the authors also provided default web snippets, which we use here. The re-partitioned version 1.1 was used. \cite{Talmor2018Repartitioning}
\item \wikihop{} \cite{welbl2017constructing}
Questions are entity-relation pairs from Freebase, and are not phrased in natural language. Multiple Wikipedia paragraphs are given as context, and the dataset was constructed such that multi-hop reasoning is needed for answering the question.
\item \comqa{} \cite{abujabal2018comqa}:
Questions are real user questions from the WikiAnswers community QA platform. No contexts are provided, and thus we augment the questions with web snippets retrieved from Google search engine.
\item \drop{} \cite{dua2019drop}:
Contexts are Wikipedia paragraphs and questions are authored by crowdsourcing workers. This dataset focuses on quantitative reasoning. Because most questions are not extractive, we only use the 33,573 extractive examples in the dataset (but evaluate on the entire development set).
\end{enumerate}

%% file: 03_models.tex
\section{Models}
\label{sec:models}

We carry our empirical investigation using two models. 
The first is \docqa{} \cite{clark2018simple}, and the second is based on BERT \cite{devlin2019bert}, which we term \bertqa{}. We now describe the pre-processing on the datasets, and provide a brief description of the models. We emphasize that in all our experiments we use exactly the same training procedure for all datasets, with minimal hyper-parameter tuning.

\paragraph{Pre-processing}
Examples in all datasets contain a question, text documents, and an answer.  To generate an extractive example we
(a) \emph{Split}: We define a length $L$ and split every paragraph whose length is $>L$ into chunks using a few manual rules.
(b) \emph{Sort}: We sort all chunks (paragraphs whose length is $\leq L$ or split paragraphs) by cosine similarity to the question in tf-idf space, as proposed by \newcite{clark2018simple}.
(c) \emph{Merge}: We go over the sorted list of chunks and greedily merge them to the largest possible length that is at most $L$, so that the RC model will be exposed to as much context as possible. The final context is the merged list of chunks $c = (c_1, \dots, c_{|c|})$
(d) We take the gold answer and mark all spans that match the answer.

\paragraph{\docqa{}} \cite{clark2018simple}:
A widely-used RC model, based on \textsc{BiDAF} \cite{seo2016bidaf}, that encodes the question and document with bidirectional RNNs, performs attention between the question and document, and adds self-attention on the document side. 

We run \docqa{} on each chunk $c_i$, where the input is a sequence of up to $L$($=400$) tokens represented as GloVE embeddings \cite{pennington2014glove}. The output is a distribution over the start and end positions of the predicted span, and we output the span with highest probability across all chunks. At training time, \docqa{} uses a shared-norm objective that normalizes the probability distribution over spans from all chunks. We define the gold span to be the first occurrence of the gold answer in the context $c$.

\paragraph{\bertqa{}} \cite{devlin2019bert}:
For each chunk, we apply the standard implementation, where the input is a sequence of $L=512$ wordpiece tokens composed of the question and chunk separated by special tokens \texttt{[CLS] <question> [SEP] <chunk> [SEP]}. A linear layer with softmax over the top-layer \texttt{[CLS]} outputs a distribution over start and end span positions.

We train over each chunk separately, back-propagating into BERT's parameters. We maximize the log-likelihood of the first occurrence of the gold answer in each chunk that contains the gold answer. At test time, we output the span with the maximal logit across all chunks.

%% file: 04_experiments.tex
\section{Controlled Experiments}
\label{sec:experiments}

We now present controlled experiments aiming to
explore generalization and transfer of models trained on a set of RC datasets to a new target. 

\subsection{Do models generalize to unseen datasets?}
\label{subsec:generalization}

We first examine generalization -- whether models trained on one dataset generalize to examples
from a new distribution. While different datasets differ substantially, there is overlap between them in terms of: (i) the language of the question, (ii) the language of the context, and (iii) the type of linguistic phenomena the dataset aims to probe. Our goal is to answer (a) do models over-fit to a particular dataset? How much does performance drop when generalizing to a new dataset? (b) Which datasets generalize better to which datasets? What properties determine generalization?

We train \docqa{} and \bertqa{} (we use BERT-base) on six large datasets (for \triviaqa{} we use \triviaqag{} and \triviaqawiki{}), taking 75K examples from each dataset to control for size. We also create \mixa{}, which contains 15K examples from the five large dataset (Using \triviaqag{} only for \triviaqa{}), resulting in another dataset of 75K examples. We evaluate performance on all datasets that the model was not trained on.

Table~\ref{tab:generalization} shows exact match (EM) performance  (does the predicted span exactly match the gold span) on the development set. The row \self{} corresponds to training and testing on the target itself, and is provided for reference (For \drop, we train on questions where the answer is a span in the context, but evaluate on the entire development set). The top part shows \docqa{}, while the bottom \bertqa{}.

At a high-level we observe three trends. 
First, models generalize poorly in this zero-shot setup: comparing \self{} to the best zero-shot number shows a performance reduction of 31.5\% on average. This confirms the finding that models over-fit to the particular dataset.
Second, \bertqa{} substantially improves generalization compared to \docqa{} owing to the power of large-scale unsupervised learning --  performance improves by 21.2\% on average. Last, \mixa{} performs almost as well as the best source dataset, reducing performance by only 3.7\% on average. Hence, training on multiple datasets results in robust generalization. We further investigate 
training on multiple datasets in \S\ref{subsec:transfer} and \S\ref{sec:full_experiments}.

\begin{table*}[t]
\centering
\resizebox{1.0\textwidth}{!}{
\begin{tabular}{lrrrrr:rrrrrrr}
\toprule
{} &   \cq &  \cwq &  \comqa &  \wikihop &  \drop &  \squad &  \newsqa &  \searchqa &  \triviaqag &  \triviaqawiki &  \hotpotqa \\
\midrule
\squad          &  18.0 &  10.1 &    16.1 &       4.2 &    2.4 &       - &     \textbf{23.4} &        9.5 &        32.0 &           20.9 &        \textbf{7.6} \\
\newsqa         &  14.9 &   8.2 &    13.5 &       4.8 &    3.0 &    \textbf{41.9} &        - &        7.7 &        25.3 &           19.9 &        5.3 \\
\searchqa       &  29.2 &  16.1 &    24.6 &       8.1 &    2.3 &    17.4 &     10.8 &          - &        \textbf{50.3} &           \textbf{28.9} &        4.5 \\
\triviaqag      &  \textbf{30.3} &  \textbf{17.8} &    \textbf{29.4} &       \textbf{9.2} &    3.0 &    30.2 &     15.5 &       \textbf{38.5} &           - &           - &        7.2 \\
\triviaqawiki   &  24.6 &  14.5 &    17.9 &       8.4 &    2.9 &    24.8 &     15.0 &       20.5 &        - &              - &        6.5 \\
\hotpotqa       &  24.6 &  14.9 &    21.2 &       8.5 &    \textbf{7.7} &    38.3 &     16.9 &       13.5 &        36.8 &           26.0 &          - \\
\hdashline
\mixa           &  32.8 &  17.9 &    26.7 &       7.4 &    4.3 &       - &        - &          - &           - &              - &          - \\
\self           &  24.1 &  24.9 &    45.2 &      41.7 &   15.6 &    68.0 &     36.5 &       51.3 &        58.9 &           41.6 &       22.5 \\
\midrule
\squad          &  23.6 &  12.0 &    20.0 &       4.6 &    5.5 &       - &     \textbf{31.8} &        8.4 &        37.8 &           33.4 &       \textbf{11.8} \\
\newsqa         &  24.1 &  12.4 &    18.9 &       7.1 &    4.4 &    \textbf{60.4} &        - &       10.1 &        37.6 &           28.4 &        8.0 \\
\searchqa       &  30.3 &  18.5 &    25.8 &      12.4 &    2.8 &    23.3 &     12.7 &          - &        \textbf{53.2} &           \textbf{35.4} &        5.2 \\
\triviaqag      &  \textbf{35.4} &  \textbf{19.7} &    \textbf{28.6} &       6.3 &    3.6 &    36.3 &     18.8 &       \textbf{39.2} &           - &              - &        8.8 \\
\triviaqawiki   &  30.3 &  16.5 &    23.6 &      \textbf{12.6} &    5.1 &    35.5 &     19.4 &       27.8 &           - &              - &        8.7 \\
\hotpotqa       &  27.7 &  15.5 &    22.1 &      10.2 &   \textbf{9.1} &    54.5 &     25.6 &       19.6 &        37.3 &           34.9 &          - \\
\hdashline
\mixa           &  34.0 &  18.2 &    30.9 &      11.7 &    8.6 &       - &        - &          - &           - &              - &          - \\
\self           &  30.8 &  27.1 &    51.6 &      52.9 &   17.9 &    78.0 &     46.0 &       52.2 &        60.7 &           50.1 &       24.2 \\
\end{tabular}}
\caption{Exact match on the development set for all datasets in a zero-shot training setup (no training on the target dataset). The top of the table shows results for \docqa{}, while the bottom for \bertqa{}. Rows correspond to the training dataset and columns to the evaluated dataset. Large datasets are on the right side, and small datasets on the left side, see text for details of all rows. Datasets used for training were not evaluated.  In \mixa{} these comprise all large datasets, and thus these cases are marked by ``-"} 
\label{tab:generalization}
\end{table*}

Taking a closer look, the pair \searchqa{} and \triviaqag{} exhibits the smallest performance drop, since both use trivia questions and web snippets. \squad{} and \newsqa{} also generalize  well (especially with \bertqa{}), probably because they contain questions on a single document, focusing on predicate-argument structure. While \hotpotqa{} and \wikihop{} both examine multi-hop reasoning over Wikipedia, performance dramatically drops from \hotpotqa{} to \wikihop{}. This is due to the difference in the language of the questions (\wikihop{} questions are synthetic). 
The best generalization to \drop{} is from \hotpotqa{}, since both require multi-hop reasoning.
Performance on \drop{} is overall low, showing that our models struggle with quantitative reasoning.

For the small datasets, \comqa{}, \cq{}, and \cwq{}, generalization is best with \triviaqag{}, as the contexts in these datasets are web snippets.
For \cq{}, whose training set has 1,300 examples, zero-shot performance is even higher than \self{}.

Interestingly, \bertqa{} improves performance substantially compared to \docqa{} on \newsqa{}, \squad{}, \triviaqawiki{} and \wikihop{}, but only moderately on \hotpotqa{}, \searchqa{}, and \triviaqag{}. 
This hints that BERT is efficient when the context is similar to (or even \emph{part of}) its training corpus, but degrades over web snippets. This is most evident when comparing \triviaqag{} to \triviaqawiki{}, as the difference between them is the type of context.

\paragraph{Global structure}
To view the global structure of the datasets, we visualize them with the force-directed placement algorithm \cite{fruchterman1991graph}. The input is a set of nodes (datasets), and a set of undirected edges representing springs in a mechanical system pulling nodes towards one another. Edges specify the pulling force, and a physical simulation places the nodes in a final minimal energy state in 2D-space. 

Let $P_{ij}$ be the performance when training \bertqa{} on dataset $D_i$ and evaluating on $D_j$. Let $P_i$ be the performance when training and evaluating on $D_i$. The force between an unordered pair of datasets is $F(D_1, D_2) = \frac{P_{12}}{P_2} + \frac{P_{21}}{P_1}$ when we train and evaluate in both directions, and $F(D_1, D_2) = \frac{2 \cdot P_{12}}{P_2}$, if we train on $D_1$ and evaluate on $D_2$ only.

\begin{figure}[t]
    \center
  \includegraphics[width=0.45\textwidth]{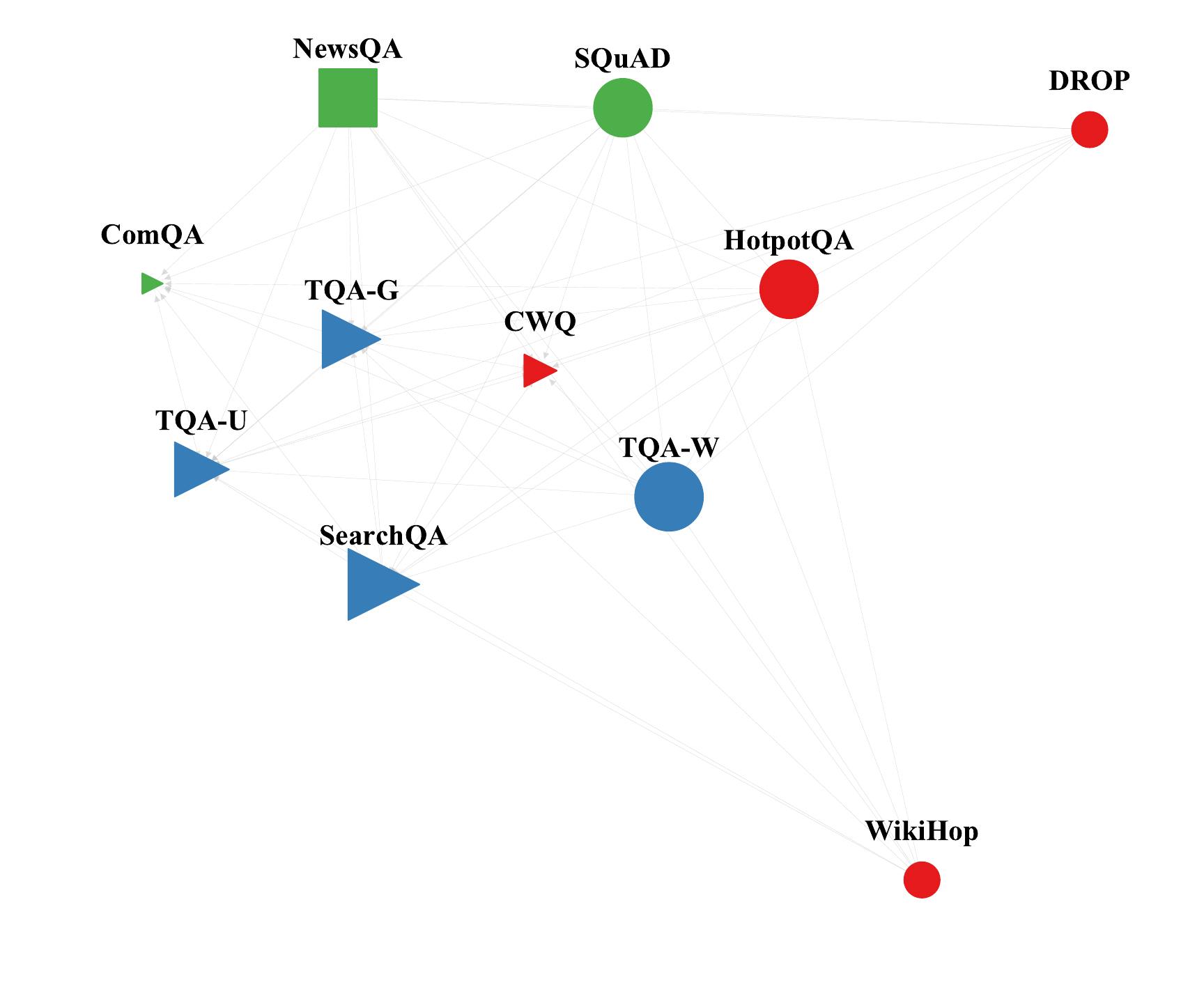}
  \caption{A 2D-visualization of the similarity between different datasets using the force-directed placement algorithm. We mark datasets that use web snippets as context with triangles, Wikipedia with circles, and Newswire with squares. We color multi-hop reasoning datasets in red, trivia datasets in blue, and factoid RC datasets in green.
  }~\label{fig:spring-graph}
\end{figure}

Figure~\ref{fig:spring-graph} shows this visualization, where we observe that datasets cluster naturally according to shape and color.
Focusing on the context, datasets with web snippets are clustered (triangles), while datasets that use Wikipedia are also near one another (circles). Considering the question language, \triviaqag{}, \searchqa{}, and \triviaqaunfilt{} are very close (blue triangles), as all contain trivia questions over web snippets. \drop{}, \hotpotqa{}, \newsqa{} and \squad{} generate questions with crowd workers, and all are at the top of the figure. \wikihop{} uses synthetic questions that prevent generalization, and is far from other datasets -- however this gap will be closed during transfer learning (\S\ref{subsec:transfer}). \drop{} is far from all datasets because it requires quantitative reasoning that is missing from other datasets. However, it is relatively close to \hotpotqa{} and \wikihop{}, which target multi-hop reasoning. \drop{} is also close to \squad{}, as both have similar contexts and question language, but the linguistic phenomena they target differ.

\begin{table}[t]
\begin{center}
\large{
\resizebox{1.0\columnwidth}{!}{
\begin{tabular}{lrrrrr}
\toprule
{} &   \cq &  \cwq &  \comqa &  \wikihop &  \drop \\
\midrule
\mixs           &  30.9 &  17.7 &    28.4 &      12.3 &    6.3 \\
\mixa           &  34.0 &  18.2 &    30.9 &      11.7 &    8.6 \\
\mixb           &  35.0 &  17.6 &    30.0 &      12.4 &    9.1 \\
\mixc           &  35.6 &  20.2 &    31.1 &      11.9 &   11.0 \\
\mixd           &  37.6 &  18.8 &    31.5 &      13.5 &   10.4 \\
\mixe           &  36.1 &  20.7 &    31.3 &      13.3 &   11.3 \\
\bottomrule
\end{tabular}}}
\end{center}
\caption{Exact match on the development set of all small datasets, as we increase the number of examples taken from the five large datasets (zero-shot setup).
}
\label{tab:multiqa_gen}
\end{table}

\paragraph{Does generalization improve with more data?}
So far we trained on datasets with 75K examples. 
To examine generalization as the training set size increases, we evaluate performance as the number of examples from the five large datasets grows. Table~\ref{tab:multiqa_gen} shows that generalization  improves by 26\% on average when increasing the number of examples from 37K to 375K.



\subsection{Does pre-training improve results on small datasets?}
\label{subsec:transfer}

\begin{table*}[t]
\centering
\resizebox{1.0\textwidth}{!}{
\begin{tabular}{lrrrrr:rrrrrr}
\toprule
{} &   \cq &  \cwq &  \comqa &  \wikihop &  \drop &  \squad &  \newsqa &  \searchqa &  \triviaqag &  \triviaqawiki &  \hotpotqa \\
\midrule
\squad        &  29.7 &  25.3 &    37.1 &      39.2 &   14.5 &       - &     33.3 &       39.2 &        49.2 &           34.5 &       17.8 \\
\newsqa       &  16.9 &  26.1 &    34.7 &      38.1 &   14.3 &    59.6 &        - &       41.6 &        44.2 &           33.9 &       16.5 \\
\searchqa     &  30.8 &  28.8 &    41.3 &      39.0 &   \textbf{15.0} &    57.0 &     31.4 &          - &        \textbf{57.5} &           39.6 &       \textbf{19.2} \\
\triviaqag    &  \textbf{41.5} &  \textbf{30.1} &    \textbf{42.6} &      \textbf{42.0} &   14.0 &    57.7 &     \textbf{31.8} &       \textbf{49.5} &           - &           \textbf{41.4} &       19.1 \\
\triviaqawiki &  31.3 &  27.0 &    38.0 &      41.4 &   13.3 &    57.6 &     31.7 &       44.4 &        50.7 &              - &       17.2 \\
\hotpotqa     &  40.0 &  27.7 &    39.5 &      40.4 &   14.6 &    \textbf{59.8} &     32.4 &       46.3 &        54.6 &           37.4 &          - \\
\hdashline
\mixa         &  43.1 &  27.6 &    39.1 &      38.9 &   14.5 &    59.8 &     33.0 &       47.5 &        56.4 &           40.4 &       19.2 \\
\self         &  24.1 &  24.9 &    45.2 &      41.7 &   15.6 &    56.5 &     30.0 &       35.9 &        41.2 &           27.7 &       13.8 \\
\midrule
\squad        &  36.9 &  29.0 &    52.2 &      48.2 &   \textbf{18.6} &       - &     \textbf{41.2} &       47.8 &        55.2 &           45.4 &       20.8 \\
\newsqa       &  36.9 &  29.4 &    52.2 &      48.4 &   17.8 &    \textbf{72.1} &        - &       47.4 &        55.9 &           45.2 &       20.6 \\
\searchqa     &  \textbf{40.5} &  30.0 &    53.4 &      \textbf{50.6} &   17.6 &    70.2 &     40.2 &          - &        \textbf{57.3} &           45.5 &       20.4 \\
\triviaqag    &  40.0 &  \textbf{30.6} &    53.4 &      49.5 &   17.6 &    69.9 &     \textbf{41.2} &       \textbf{50.0} &           - &           \textbf{46.2} &       20.8 \\
\triviaqawiki &  39.0 &  30.3 &    \textbf{54.0} &      50.0 &   17.3 &    71.0 &     39.2 &       48.4 &        55.7 &              - &       \textbf{20.9} \\
\hotpotqa     &  34.4 &  30.2 &    53.0 &      49.3 &   17.2 &    71.2 &     39.5 &       48.6 &        56.6 &           45.6 &          - \\
\hdashline
\mixa         &  42.6 &  30.6 &    53.3 &      50.5 &   17.9 &    71.5 &     42.1 &       48.5 &        56.6 &           46.5 &       20.4 \\
\self         &  30.8 &  27.1 &    51.6 &      52.9 &   17.1 &    70.1 &     37.9 &       46.0 &        54.4 &           41.9 &       18.9 \\
\bottomrule
\end{tabular}}
\caption{Exact match on the development set for all datasets with transfer learning. Fine-tuning is done on $\leq 15K$ examples.
The top of the table shows results for \docqa{}, while the bottom for \bertqa{}. Rows are the trained datasets and columns are the evaluated datasets for which fine-tuning was performed.  Large datasets are on the right, and small datasets are on the left side}
\label{tab:transfer}
\end{table*}

We now consider transfer learning, assuming access to a small number of examples ($\leq$15K) from a target dataset. 
We pre-train a model on a source dataset, and then fine-tune on the target.  In all models, pre-training and fine-tuning are identical and performed until no improvement is seen on the development set (early stopping).
Our goal is to analyze whether pre-training improves performance compared to training on the target alone. This is particularly interesting with \bertqa{}, as BERT already contains substantial knowledge that might deem pre-training unnecessary. 

\paragraph{How to choose the dataset to pre-train on?} 
Table~\ref{tab:transfer} shows exact match (EM) on the development set of all datasets (rows are the trained datasets and columns the evaluated datasets). Pre-training on a source RC dataset and transferring to the target improves performance by 21\% on average for \docqa{} (improving on 8 out of 11 datasets), and by 7\% on average for \bertqa{} (improving on 10 out of 11 datasets). Thus, pre-training on a related RC dataset helps even given representations from a model like \bertqa{}.

Second, \mixa{} obtains good performance in almost all setups. Performance of \mixa{} is 3\% lower than the best source RC dataset on average for \docqa{}, and 0.3\% lower for \bertqa{}. Hence, one can pre-train a single model on a mixed dataset, rather than choose the best source dataset for every target.

Third, in 4 datasets (\comqa{}, \drop{}, \hotpotqa{}, \wikihop{}) the best source dataset uses web snippets in \docqa{}, but Wikipedia in \bertqa{}. This strengthens our finding that \bertqa{} performs better given Wikipedia text.

Last, we see dramatic improvement in performance comparing to \S\ref{subsec:generalization}.
This highlights that current models over-fit to the data they are trained on, and small amounts of data from the target distribution can overcome this generalization gap. This is clearest for \wikihop{}, where synthetic questions preclude generalization, but fine-tuning improves performance from 12.6 EM to 50.5 EM. Thus, low performance was not due to a modeling issue, but rather a mismatch in the question language.

An interesting question is whether performance in the generalization setup is predictive of performance in the transfer setup. Average performance across target datasets in Table~\ref{tab:transfer}, when choosing the best source dataset from Table~\ref{tab:transfer}, is 39.3 (\docqa{}) and 43.8 (\bertqa{}). Average performance across datasets in Table~\ref{tab:transfer}, when choosing the best source dataset from Table~\ref{tab:generalization}, is 38.9 (\docqa{}) and 43.5 (\bertqa{}).
Thus, one can select a dataset to pre-train on based on generalization performance and suffer a minimal hit in accuracy, without fine-tuning on each dataset. However, training on \mixa{} also yields good results without selecting a source dataset at all.



\begin{figure*}[ht]
    \center
  \includegraphics[width=1.0\textwidth]{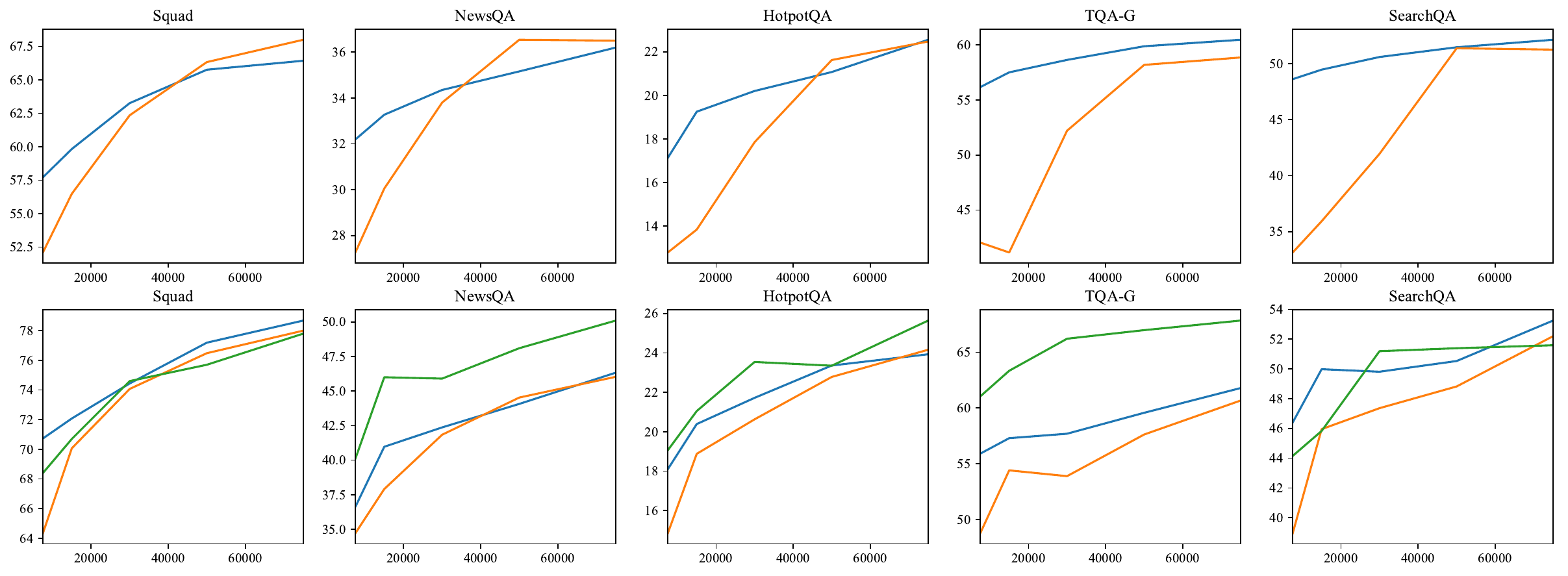}
  \caption{
  Learning curves for the five large datasets (top is \docqa{} and bottom is \bertqa{}). The x-axis corresponds to the number of examples from the target dataset, and the y-axis is EM. The orange curve refers to training on the target dataset only, and the blue curve refers to pre-training on 75K examples from the nearest source dataset and fine-tuning on the target dataset. The green curve is training on a fixed number of examples from all 5 large datasets without fine-tuning (\multiqa{}). 
 }~\label{fig:learning-curves}
\end{figure*}

\paragraph{How much target data is needed?}
We saw that with 15K training examples from the target dataset, pre-training improves performance. We now ask whether this effect maintains given a larger training set. To examine this, we measure (Figure~\ref{fig:learning-curves}) the
performance on each of the large datasets when pre-training on its nearest dataset (according to $F(\cdot, \cdot)$) for both \docqa{} (top) and \bertqa{} (bottom row). The orange curve corresponds to training on the target dataset only, while the blue curve describes pre-training on 75K examples from a source dataset, and then fine-tuning on an increasing number of examples from the target dataset.

In 5 out of 10 curves, pre-training improves performance even given access to all 75K examples from the target dataset. In the other 5, using only the target dataset is better after 30-50K examples.
To estimate the 
savings in annotation costs through pre-training, we measure  how many examples are needed, when doing pre-training, to reach 95\% of the performance obtained when training on all examples from the target dataset. We find that with pre-training we only need 49\% of the examples to reach 95\% performance, compared to 86\% without pre-training.

To further explore pre-training on multiple datasets, we plot a curve (green) for \bertqa{}, where at each point we train on a fixed number of examples from all five large datasets (no fine-tuning). We observe that more data from multiple datasets improves performance in almost all cases. In this case, we reach 95\% of the final performance using 30\% of the examples only.
We will use this observation further in \S\ref{sec:full_experiments} to reach new state-of-the-art performance on several datasets.

\subsection{Does context augmentation improve performance?}
For \triviaqa{} we have for all questions, contexts from three different sources -- Wikipedia (\triviaqawiki{}), Bing web snippets (\triviaqaunfilt{}), and Google web snippets (\triviaqag{}). Thus, we can explore whether combining the three datasets improves performance. Moreover, because questions are identical across the datasets, we can see the effect on generalization due to the context language only.

Table~\ref{tab:triviaqa} shows the results. In the first 3 rows we train on 75K examples from each dataset, and in the last we train on the combined 225K examples. First, we observe that context augmentation substantially improves performance (especially for \triviaqag{} and \triviaqawiki{}). Second, generalization is sensitive to the context type: performance substantially drops when training on one context type and evaluating on another ($60.7 \rightarrow$ 48.4 for \triviaqag{}, $53.1 \rightarrow 44.6$ for \triviaqaunfilt{}, and $50.1 \rightarrow 43.3$ for \triviaqawiki{}).

\begin{table}[t]
\begin{center}
\scriptsize{
\resizebox{1.0\columnwidth}{!}{
\begin{tabular}{lrrr}
\toprule
{} &  \triviaqag &  \triviaqaunfilt &  \triviaqawiki \\
\midrule
\triviaqag                                 &        60.7 &             53.6 &           43.3 \\
\triviaqaunfilt                            &        57.2 &             53.1 &           39.9 \\
\triviaqawiki                              &        48.4 &             44.6 &           50.1 \\
\triviaqamix &        67.7 &             54.4 &           54.7 \\
\bottomrule
\end{tabular}}}
\end{center}
\caption{EM on the development set, where
each row uses the same question with a different context, and \triviaqamix{} is a union of the other 3 datasets.}
\label{tab:triviaqa}
\end{table}

%% file: 05_full_experiments.tex
\section{\multiqa{}}
\label{sec:full_experiments}

\begin{table*}[t]
\centering
\scriptsize{
\resizebox{0.9\textwidth}{!}{
\renewcommand{\arraystretch}{1.1}
\begin{tabular}{l|cc|cc|cc|cc}
& \multicolumn{2}{c|}{BERT-large Dev.} & \multicolumn{2}{c|}{\multiqa{} Dev.} & \multicolumn{2}{c|}{\multiqa{} Test} & \multicolumn{2}{c}{\textsc{SOTA}\footnotemark[1]} \\ 
\cline{1-9}
Dataset                       & EM            & tok. F1           & EM        & tok. F1         & EM        & tok. F1          & EM        & tok. F1         \\
\cline{1-9}
\newsqa                     & 51.5          & 66.2         & 53.9      & 68.2       & \textbf{52.3}      & \textbf{67.4}        & 53.1    &  66.3        \\
\searchqa                   & 59.2          & 66.4         & 60.7      & 67.1       & \textbf{59.0}      & \textbf{65.1}        & 58.8      & 64.5        \\
\triviaqaunfilt             & 56.8          & 62.6         & 58.4      & 64.3       & -         & -           & 52.0\footnotemark[2]      & 61.7\footnotemark[2]   \\
\cwq                        & 30.8          & -            & 35.4      &  -          & \textbf{34.9}      & -           & 34.2      & -           \\
\hotpotqa                   & 27.9          & 37.7         & 30.6      & 40.3       & \textbf{30.7}         & \textbf{40.2}           & 37.1\footnotemark[2]	  & 48.9\footnotemark[2]         \\ 
\bottomrule
\end{tabular}}}
\caption{Results for datasets where the official evaluation metric is EM and token F$_1$.
The \cwq{} evaluation script provides only the EM mertic.  We did not find a public evaluation script for the hidden test set of \triviaqaunfilt{}.}
\label{tab:main_res}
\end{table*}

\begin{table*}[t]
\centering
\scriptsize{
\resizebox{0.9\textwidth}{!}{
\renewcommand{\arraystretch}{1.1}
\begin{tabular}{l|ccc|ccc|ccc|ccc}
& \multicolumn{3}{c|}{BERT-large Dev.} & \multicolumn{3}{c|}{\multiqa{} Dev.} & \multicolumn{3}{c|}{\multiqa{} Test} & \multicolumn{3}{c}{\textsc{SOTA}} \\ 
\cline{1-13}
Dataset                   & Prec. & Rec.  & F1   & Prec. & Rec.  & F1    & Prec. & Rec. & F1     & Prec.  & Rec. & F1       \\ 
\cline{1-13}
\comqa                  & 45.8  & 42.0  & 42.9 &  51.9 & 47.2  & 48.2  & \textbf{44.4}  & \textbf{40.0} & \textbf{40.8}   & 21.2   & 38.4 & 22.4     \\
\cq                     &   -    &     -  &  32.8  &  -    &  - &       46.6       &  -   & -   & \textbf{42.4}     & -   & - & 39.7\footnotemark[2]      \\
\bottomrule
\end{tabular}}}
\caption{Results for datasets where the evaluation metric is average recall/precision/F$_1$. 
\cq{} evaluates with F$_1$ only.
} 
\label{tab:main_res_answer_lists}
\end{table*}

We now present \multiqa{}, a BERT-based model, trained on multiple RC datasets, that obtains new state-of-the-art results on several datasets.

\paragraph{Does training on multiple datasets improve \bertqa{}?}
\multiqa{} trains \bertqa{} on the \mixe{} dataset presented above, which contains 75K examples from 5 large datasets, but uses BERT-large rather than BERT-base.
For small target datasets, we fine-tune the model 
on these datasets, since they were not observed when training on \mixe{}. For large datasets, we do not fine-tune.
We found that fine-tuning on datasets that are already part of \mixe{} does not improve performance (we assume this is due to the high-capacity of BERT-large), and thus we use \emph{one model} for all the large datasets. We train on \mixe{}, and thus our model does not use all examples in the original datasets, which contain more than 75K examples.

We use the official evaluation script for any dataset that provides one, and the \squad{} evaluation script for all other datasets. Table~\ref{tab:main_res} shows results for datasets where the evaluation metric is EM or token F$_1$ (harmonic mean of the list of tokens in the predicted vs. gold span). Table~\ref{tab:main_res_answer_lists} shows results for datasets where the  evaluation metric is average recall/precision/F$_1$ between the list of predicted answers and the list of gold answers.

We compare \multiqa{} to BERT-large, a model that does not train on \mixe{}, but only fine-tunes BERT-large on the target dataset.
We also show the state-of-the-art (\textsc{SOTA}) result for all datasets for reference.\footnote{State-of-the-are-results were found in \cite{Tay2018Advances} for \newsqa{}, in \newcite{lin2018denoising}, for \searchqa{}, in \newcite{das2019multi} for \triviaqaunfilt{}, in \cite{talmor2018repart} for \cwq{}, in \newcite{Ding2019Cognitive} for \hotpotqa{}, in  \cite{abujabal2018comqa} for \comqa{}, and in \newcite{bao2016constraint} for \cq{}.
}

\multiqa{} improves state-of-the-art performance on \sotasnum datasets, although it does not even train on all examples in the large datasets.\footnote{We compare only to models for which we found a publication.
For \triviaqaunfilt{}, Figure 4 in \newcite{clark2018simple} shows roughly 67 F$_1$ on the development set, but no exact number.
For \cq{} we compare against SOTA achieved on the web snippets context. On the Freebase context SOTA is 42.8 F$_1$.
\cite{Luo12018Empirical}}
\multiqa{} improves performance compared to BERT-large in all cases.
This improvement is especially noticeable in small datasets such as \comqa{}, \cwq{}, and \cq{}. Moreover, in \newsqa{}, \multiqa{} 
surpasses human performance as measured by the creators of those datasets.
(46.5 EM, 69.4 F1)  \cite{trischler2017newsqa}), improving upon previous state-of-the-art by a large margin.

To conclude, \multiqa{} is able to improve state-of-the-art performance on multiple datasets. Our results suggest that in many NLU tasks the size of the dataset is the main bottleneck rather than the model itself.





\paragraph{
Does training on multiple datasets improve resiliency against adversarial attacks?}
Finally, we evaluated \multiqa{} on the
adversarial \squad{}  \cite{jia2017adversarial}, where a misleading sentence is appended to each context (\textsc{AddSent} variant). \multiqa{} obtained 66.7 EM and 73.1 F$_1$, outperforming BERT-large (60.4EM, 66.3F1) by a significant margin, and also substantially improving state-of-the-art results (56.0 EM, 61.3 F$_1$, \cite{hu2018attention} and  52.1 EM, 62.7 F$_1$, \cite{wang2018Multi}).

%% file: 06_related.tex
\section{Related Work}
Prior work has shown that RC performance can be improved by training on a large dataset and transferring to a smaller one, but at a small scale \cite{min2017transfer,chung2018supervised}. \newcite{sun2018improving} has recently shown this in a larger experiment for multi-choice questions, where they first fine-tuned BERT on \textsc{RACE} \cite{lai2017race} and then fine-tuned on several smaller datasets.

Interest in learning general-purpose representations 
for natural language through unsupervised, multi-task and transfer learning has been sky-rocketing lately \cite{peters2018elmo,radford2018improving,mccann2018natural,chronopoulou2019transfer,phang2018sentence,wang2019glue}.
In parallel to our work, studies that focus on generalization have appeared on publication servers, empirically studying generalization to multiple tasks
\cite{yogatama2019learning,liu2019multi}. Our work is part of this research thread on generalization in natural langauge understanding, focusing on reading comprehension, which we view as an important and broad language understanding task.

%% file: 07_conclusions.tex
\section{Conclusions}
\label{sec:conclusions}

In this work we performed a thorough empirical investigation of generalization and transfer over 10 RC datasets. 
We characterized the factors affecting generalization and obtained several state-of-the-art results by training on 375K examples from 5 RC datasets. We open source our infrastructure for easily performing experiments on multiple RC datasets, for the benefit of the community.

We highlight several practical take-aways:
\begin{itemize}[topsep=0pt, itemsep=0pt, leftmargin=0in, parsep=0pt]
    \item Pre-training on multiple source RC datasets consistently improves performance on a target RC dataset ,
    even in the presence of BERT representations. It also leads to substantial reduction in the number of necessary training examples for a fixed performance.
    \item Training the high-capacity BERT-large representations over multiple RC datasets leads to good performance on all of the trained datasets
    without having to fine-tune on each dataset separately.
    \item BERT representations improve generalization, but their effect is moderate when the source of the context is web snippets compared to Wikipedia and newswire.
    \item Performance over an RC dataset can be improved by retrieving web snippets for all questions and adding them as examples (context augmentation).
\end{itemize}

%% file: acl2019.bbl
\begin{thebibliography}{46}
\expandafter\ifx\csname natexlab\endcsname\relax\def\natexlab#1{#1}\fi

\bibitem[{Abujabal et~al.(2018)Abujabal, Roy, Yahya, and
  Weikum}]{abujabal2018comqa}
A.~Abujabal, R.~S. Roy, M.~Yahya, and G.~Weikum. 2018.
\newblock Comqa: A community-sourced dataset for complex factoid question
  answering with paraphrase clusters.
\newblock \emph{arXiv preprint arXiv:1809.09528}.

\bibitem[{Bao et~al.(2016)Bao, Duan, Yan, Zhou, and Zhao}]{bao2016constraint}
J.~Bao, N.~Duan, Z.~Yan, M.~Zhou, and T.~Zhao. 2016.
\newblock Constraint-based question answering with knowledge graph.
\newblock In \emph{International Conference on Computational Linguistics
  (COLING)}.

\bibitem[{Berant et~al.(2014)Berant, Srikumar, Chen, Linden, Harding, Huang,
  Clark, and Manning}]{berant2014biological}
J.~Berant, V.~Srikumar, P.~Chen, A.~V. Linden, B.~Harding, B.~Huang, P.~Clark,
  and C.~D. Manning. 2014.
\newblock Modeling biological processes for reading comprehension.
\newblock In \emph{Empirical Methods in Natural Language Processing (EMNLP)}.

\bibitem[{Bollacker et~al.(2008)Bollacker, Evans, Paritosh, Sturge, and
  Taylor}]{bollacker2008freebase}
K.~Bollacker, C.~Evans, P.~Paritosh, T.~Sturge, and J.~Taylor. 2008.
\newblock {F}reebase: a collaboratively created graph database for structuring
  human knowledge.
\newblock In \emph{International Conference on Management of Data (SIGMOD)},
  pages 1247--1250.

\bibitem[{Choi et~al.(2018)Choi, He, Iyyer, Yatskar, Yih, Choi, Liang, and
  Zettlemoyer}]{choi2018quac}
E.~Choi, H.~He, M.~Iyyer, M.~Yatskar, W.~Yih, Y.~Choi, P.~Liang, and
  L.~Zettlemoyer. 2018.
\newblock Quac: Question answering in context.
\newblock In \emph{Empirical Methods in Natural Language Processing (EMNLP)}.

\bibitem[{Chronopoulou et~al.(2019)Chronopoulou, Baziotis, and
  Potamianos}]{chronopoulou2019transfer}
A.~Chronopoulou, C.~Baziotis, and A.~Potamianos. 2019.
\newblock An embarrassingly simple approach for transfer learning from
  pretrained language models.
\newblock \emph{arXiv preprint arXiv:1902.10547}.

\bibitem[{Chung et~al.(2018)Chung, Lee, and Glass}]{chung2018supervised}
Y.~Chung, H.~Lee, and J.~Glass. 2018.
\newblock Supervised and unsupervised transfer learning for question answering.
\newblock In \emph{North American Association for Computational Linguistics
  (NAACL)}.

\bibitem[{Clark and Gardner(2018)}]{clark2018simple}
C.~Clark and M.~Gardner. 2018.
\newblock Simple and effective multi-paragraph reading comprehension.
\newblock In \emph{Association for Computational Linguistics (ACL)}.

\bibitem[{Das et~al.(2019)Das, Dhuliawala, Zaheer, and McCallum}]{das2019multi}
R.~Das, S.~Dhuliawala, M.~Zaheer, and A.~McCallum. 2019.
\newblock Multi-step retriever-reader interaction for scalable open-domain
  question answering.
\newblock In \emph{International Conference on Learning Representations
  (ICLR)}.

\bibitem[{Devlin et~al.(2019)Devlin, Chang, Lee, and
  Toutanova}]{devlin2019bert}
J.~Devlin, M.~Chang, K.~Lee, and K.~Toutanova. 2019.
\newblock Bert: Pre-training of deep bidirectional transformers for language
  understanding.
\newblock In \emph{North American Association for Computational Linguistics
  (NAACL)}.

\bibitem[{Ding et~al.(2019)Ding, Zhou, Chen, Yang, and
  Tang}]{Ding2019Cognitive}
M.~Ding, C.~Zhou, Q.~Chen, H.~Yang, and J.~Tang. 2019.
\newblock Cognitive graph for multi-hop reading comprehension at scale.
\newblock In \emph{Association for Computational Linguistics (ACL)}.

\bibitem[{Dua et~al.(2019)Dua, Wang, Dasigi, Stanovsky, Singh, and
  Gardner}]{dua2019drop}
D.~Dua, Y.~Wang, P.~Dasigi, G.~Stanovsky, S.~Singh, and M.~Gardner. 2019.
\newblock Drop: A reading comprehension benchmark requiring discrete reasoning
  over paragraphs.
\newblock In \emph{North American Association for Computational Linguistics
  (NAACL)}.

\bibitem[{Dunn et~al.(2017)Dunn, , Sagun, Higgins, Guney, Cirik, and
  Cho}]{dunn2017searchqa}
M.~Dunn, , L.~Sagun, M.~Higgins, U.~Guney, V.~Cirik, and K.~Cho. 2017.
\newblock {SearchQA}: A new {Q}\&{A} dataset augmented with context from a
  search engine.
\newblock \emph{arXiv}.

\bibitem[{Fruchterman and Reingold(1991)}]{fruchterman1991graph}
T.~M. Fruchterman and E.~M. Reingold. 1991.
\newblock Graph drawing by force-directed placement.
\newblock \emph{Software: Practice and experience}, 21(11):1129--1164.

\bibitem[{Gardner et~al.(2018)Gardner, Grus, Neumann, Tafjord, Dasigi, Liu,
  Peters, Schmitz, and Zettlemoyer}]{gardner2018allennlp}
M.~Gardner, J.~Grus, M.~Neumann, O.~Tafjord, P.~Dasigi, N.~Liu, M.~Peters,
  M.~Schmitz, and L.~Zettlemoyer. 2018.
\newblock {AllenNLP}: A deep semantic natural language processing platform.
\newblock \emph{arXiv preprint arXiv:1803.07640}.

\bibitem[{Hermann et~al.(2015)Hermann, Kočiský, Grefenstette, Espeholt, Kay,
  Suleyman, and Blunsom}]{hermann2015read}
K.~M. Hermann, T.~Kočiský, E.~Grefenstette, L.~Espeholt, W.~Kay, M.~Suleyman,
  and P.~Blunsom. 2015.
\newblock Teaching machines to read and comprehend.
\newblock In \emph{Advances in Neural Information Processing Systems
  (NeurIPS)}.

\bibitem[{Hu et~al.(2018)Hu, Peng, Wei, Huang, Li, Yang, and
  Zhou}]{hu2018attention}
M.~Hu, Y.~Peng, F.~Wei, Z.~Huang, D.~Li, N.~Yang, and M.~Zhou. 2018.
\newblock Attention-guided answer distillation for machine reading
  comprehension.
\newblock In \emph{Empirical Methods in Natural Language Processing (EMNLP)}.

\bibitem[{Jia and Liang(2017)}]{jia2017adversarial}
R.~Jia and P.~Liang. 2017.
\newblock Adversarial examples for evaluating reading comprehension systems.
\newblock In \emph{Empirical Methods in Natural Language Processing (EMNLP)}.

\bibitem[{Joshi et~al.(2017)Joshi, Choi, Weld, and
  Zettlemoyer}]{joshi2017triviaqa}
M.~Joshi, E.~Choi, D.~Weld, and L.~Zettlemoyer. 2017.
\newblock {TriviaQA}: A large scale distantly supervised challenge dataset for
  reading comprehension.
\newblock In \emph{Association for Computational Linguistics (ACL)}.

\bibitem[{Lai et~al.(2017)Lai, Xie, Liu, Yang, and Hovy}]{lai2017race}
G.~Lai, Q.~Xie, H.~Liu, Y.~Yang, and E.~Hovy. 2017.
\newblock Race: Large-scale reading comprehension dataset from examinations.
\newblock \emph{arXiv preprint arXiv:1704.04683}.

\bibitem[{Lin et~al.(2018)Lin, Ji, Liu, and Sun}]{lin2018denoising}
Y.~Lin, H.~Ji, Z.~Liu, and M.~Sun. 2018.
\newblock Denoising distantly supervised open-domain question answering.
\newblock In \emph{Association for Computational Linguistics (ACL)}, volume~1,
  pages 1736--1745.

\bibitem[{Liu et~al.(2019)Liu, He, Chen, and Gao}]{liu2019multi}
X.~Liu, P.~He, W.~Chen, and J.~Gao. 2019.
\newblock Multi-task deep neural networks for natural language understanding.
\newblock \emph{arXiv preprint arXiv:1901.11504}.

\bibitem[{Luo1 et~al.(2018)Luo1, Lin1, X., Kenny, and
  Q.Zhu1}]{Luo12018Empirical}
K.~Luo1, F.~Lin1, X., L.~Kenny, and Q.Zhu1. 2018.
\newblock Knowledge base question answering via encoding of complex query
  graphs.
\newblock In \emph{Empirical Methods in Natural Language Processing (EMNLP)}.

\bibitem[{McCann et~al.(2018)McCann, Keskar, Xiong, and
  Socher}]{mccann2018natural}
B.~McCann, N.~S. Keskar, C.~Xiong, and R.~Socher. 2018.
\newblock The natural language decathlon: Multitask learning as question
  answering.
\newblock \emph{arXiv preprint arXiv:1806.08730}.

\bibitem[{Min et~al.(2017)Min, Seo, and Hajishirzi}]{min2017transfer}
S.~Min, M.~Seo, and H.~Hajishirzi. 2017.
\newblock Question answering through transfer learning from large fine-grained
  supervision data.
\newblock In \emph{Association for Computational Linguistics (ACL)}.

\bibitem[{Nguyen et~al.(2016)Nguyen, Rosenberg, Song, Gao, Tiwary, Majumder,
  and Deng}]{nguyen2016ms}
T.~Nguyen, M.~Rosenberg, X.~Song, J.~Gao, S.~Tiwary, R.~Majumder, and L.~Deng.
  2016.
\newblock {MS MARCO}: A human generated machine reading comprehension dataset.
\newblock In \emph{Workshop on Cognitive Computing at NIPS}.

\bibitem[{Pennington et~al.(2014)Pennington, Socher, and
  Manning}]{pennington2014glove}
J.~Pennington, R.~Socher, and C.~D. Manning. 2014.
\newblock Glo{V}e: Global vectors for word representation.
\newblock In \emph{Empirical Methods in Natural Language Processing (EMNLP)},
  pages 1532--1543.

\bibitem[{Peters et~al.(2018)Peters, Neumann, Iyyer, Gardner, Clark, Lee, and
  Zettlemoyer}]{peters2018elmo}
M.~E. Peters, M.~Neumann, M.~Iyyer, M.~Gardner, C.~Clark, K.~Lee, and
  L.~Zettlemoyer. 2018.
\newblock Deep contextualized word representations.
\newblock In \emph{North American Association for Computational Linguistics
  (NAACL)}.

\bibitem[{Phang et~al.(2018)Phang, Fevry, and Bowman}]{phang2018sentence}
J.~Phang, T.~Fevry, and S.~R. Bowman. 2018.
\newblock Sentence encoders on stilts: Supplementary training on intermediate
  labeled-data tasks.
\newblock \emph{arXiv preprint arXiv:1811.01088}.

\bibitem[{Radford et~al.(2018)Radford, Narasimhan, Salimans, and
  Sutskever}]{radford2018improving}
A.~Radford, K.~Narasimhan, T.~Salimans, and I.~Sutskever. 2018.
\newblock Improving language understanding by generative pre-training.
\newblock Technical report, OpenAI.

\bibitem[{Rajpurkar et~al.(2016)Rajpurkar, Zhang, Lopyrev, and
  Liang}]{rajpurkar2016squad}
P.~Rajpurkar, J.~Zhang, K.~Lopyrev, and P.~Liang. 2016.
\newblock {SQuAD}: 100,000+ questions for machine comprehension of text.
\newblock In \emph{Empirical Methods in Natural Language Processing (EMNLP)}.

\bibitem[{Reddy et~al.(2018)Reddy, Chen, and Manning}]{reddy2018coqa}
S.~Reddy, D.~Chen, and C.~D. Manning. 2018.
\newblock Coqa: A conversational question answering challenge.
\newblock \emph{arXiv preprint arXiv:1808.07042}.

\bibitem[{Richardson et~al.(2013)Richardson, Burges, and
  Renshaw}]{richardson2013mctest}
M.~Richardson, C.~J. Burges, and E.~Renshaw. 2013.
\newblock Mctest: A challenge dataset for the open-domain machine comprehension
  of text.
\newblock In \emph{Empirical Methods in Natural Language Processing (EMNLP)},
  pages 193--203.

\bibitem[{Seo et~al.(2016)Seo, Kembhavi, Farhadi, and
  Hajishirzi}]{seo2016bidaf}
M.~Seo, A.~Kembhavi, A.~Farhadi, and H.~Hajishirzi. 2016.
\newblock Bidirectional attention flow for machine comprehension.
\newblock \emph{arXiv}.

\bibitem[{Sun et~al.(2018)Sun, Yu, Yu, and Cardie}]{sun2018improving}
K.~Sun, D.~Yu, D.~Yu, and C.~Cardie. 2018.
\newblock Improving machine reading comprehension with general reading
  strategies.
\newblock \emph{arXiv preprint arXiv:1810.13441}.

\bibitem[{Talmor and Berant(2018{\natexlab{a}})}]{Talmor2018Repartitioning}
A.~Talmor and J.~Berant. 2018{\natexlab{a}}.
\newblock Repartitioning of the complexwebquestions dataset.
\newblock \emph{arXiv preprint arXiv:1807.09623}.

\bibitem[{Talmor and Berant(2018{\natexlab{b}})}]{talmor2018repart}
A.~Talmor and J.~Berant. 2018{\natexlab{b}}.
\newblock Repartitioning of the complexwebquestions dataset.
\newblock \emph{arXiv preprint arXiv:1807.09623}.

\bibitem[{Talmor and Berant(2018{\natexlab{c}})}]{talmor2018web}
A.~Talmor and J.~Berant. 2018{\natexlab{c}}.
\newblock The web as knowledge-base for answering complex questions.
\newblock In \emph{North American Association for Computational Linguistics
  (NAACL)}.

\bibitem[{Talmor et~al.(2017)Talmor, Geva, and Berant}]{talmor2017evaluating}
A.~Talmor, M.~Geva, and J.~Berant. 2017.
\newblock Evaluating semantic parsing against a simple web-based question
  answering model.
\newblock In \emph{*SEM}.

\bibitem[{Tay et~al.(2018)Tay, Tuan, Hui, and Su}]{Tay2018Advances}
Y.~Tay, L.~Tuan, S.~Hui, and J.~Su. 2018.
\newblock Densely connected attention propagation for reading comprehension.
\newblock In \emph{Advances in Neural Information Processing Systems
  (NeurIPS)}.

\bibitem[{Trischler et~al.(2017)Trischler, Wang, Yuan, Harris, Sordoni,
  Bachman, and Suleman}]{trischler2017newsqa}
A.~Trischler, T.~Wang, X.~Yuan, J.~Harris, A.~Sordoni, P.~Bachman, and
  K.~Suleman. 2017.
\newblock {NewsQA}: A machine comprehension dataset.
\newblock In \emph{Workshop on Representation Learning for NLP}.

\bibitem[{Wang et~al.(2019)Wang, Singh, Michael, Hill, Levy, and
  Bowman}]{wang2019glue}
A.~Wang, A.~Singh, J.~Michael, F.~Hill, O.~Levy, and S.~R. Bowman. 2019.
\newblock Glue: A multi-task benchmark and analysis platform for natural
  language understanding.
\newblock In \emph{International Conference on Learning Representations
  (ICLR)}.

\bibitem[{Wang et~al.(2018)Wang, Yan, and Wu}]{wang2018Multi}
W.~Wang, M.~Yan, and C.~Wu. 2018.
\newblock Multi-granularity hierarchical attention fusion networks for reading
  comprehension and question answering.
\newblock In \emph{Association for Computational Linguistics (ACL)}.

\bibitem[{Welbl et~al.(2017)Welbl, Stenetorp, and
  Riedel}]{welbl2017constructing}
J.~Welbl, P.~Stenetorp, and S.~Riedel. 2017.
\newblock Constructing datasets for multi-hop reading comprehension across
  documents.
\newblock \emph{arXiv preprint arXiv:1710.06481}.

\bibitem[{Yang et~al.(2018)Yang, Qi, Zhang, Bengio, Cohen, Salakhutdinov, and
  Manning}]{yang2018hotpotqa}
Z.~Yang, P.~Qi, S.~Zhang, Y.~Bengio, W.~W. Cohen, R.~Salakhutdinov, and C.~D.
  Manning. 2018.
\newblock {HotpotQA}: A dataset for diverse, explainable multi-hop question
  answering.
\newblock In \emph{Empirical Methods in Natural Language Processing (EMNLP)}.

\bibitem[{Yogatama et~al.(2019)Yogatama, de~M.~d'Autume, Connor, Kocisky,
  Chrzanowski, Kong, Lazaridou, Ling, Yu, Dyer et~al.}]{yogatama2019learning}
D.~Yogatama, C.~de~M.~d'Autume, J.~Connor, T.~Kocisky, M.~Chrzanowski, L.~Kong,
  A.~Lazaridou, W.~Ling, L.~Yu, C.~Dyer, et~al. 2019.
\newblock Learning and evaluating general linguistic intelligence.
\newblock \emph{arXiv preprint arXiv:1901.11373}.

\end{thebibliography}
